\documentclass[multiauthors,onecolumn]{LaTeXclass/cohere}

\usepackage{ragged2e}
\usepackage{lipsum}
\usepackage{pdfpages}
\usepackage{colortbl}
\usepackage{tabulary}
\usepackage{longtable}
\usepackage{array}
\usepackage{placeins}
\usepackage{tablefootnote}
\usepackage{tcolorbox}
\usepackage{wrapfig}
\usepackage{breqn} 

\usepackage{pifont}
\definecolor{deeppurple}{HTML}{9e02f7}
\definecolor{forestgreen}{HTML}{2e7d43}
\usepackage{multirow}
\usepackage{tabularx}

\usepackage{enumitem}
\usepackage{arydshln}
\usepackage{amsmath}
\usepackage{amsfonts}
\usepackage{cleveref}
\usepackage{wrapfig}
\usepackage{tcolorbox}
\usepackage{mdframed}

\usepackage{fancyvrb}
\usepackage{fvextra}

\usepackage[utf8]{inputenc}

\usepackage{arydshln}

\usepackage{xspace} 
\usepackage{makecell} 
\usepackage{multirow}

\usepackage{nicematrix}
\usepackage{comment}
\usepackage{amssymb}

\setcitestyle{number,square}

\title{Treasure Hunt: Real-time Targeting of the Long Tail using Training-Time Markers}

\author{name={Daniel D'souza\fa},affiliation={1}}
\author{name={Julia Kreutzer},affiliation={1}}
\author{name={Adrien Morisot},affiliation={2}}
\author{name={Ahmet Üstün\psa},affiliation={1}}
\author{name={Sara Hooker\psa},affiliation={1}}
\affiliations{
    \item[1] Cohere Labs
    \item[2] Cohere
}

\corresponding[*]{\{\url{danieldsouza}, \url{ahmet}, \url{sarahooker}\}\url{{@cohere.com}}}

\abstract{
\justifying
One of the most profound challenges of modern machine learning is performing well on the long-tail of rare and underrepresented features. Large general-purpose models are trained for many tasks, but work best on high-frequency use cases. After training, it is hard to adapt a model to perform well on specific use cases underrepresented in the training corpus. Relying on prompt engineering or few-shot examples to maximize the output quality on a particular test case can be frustrating, as models can be highly sensitive to small changes, react in unpredicted ways or rely on a fixed system prompt for maintaining performance. 
In this work, we ask: \textit{Can we optimize our training protocols to both improve controllability and performance on underrepresented use cases at inference time?} 
We revisit the divide between training and inference techniques to improve long-tail performance while providing users with a set of control levers the model is trained to be responsive to. We create a detailed taxonomy of data characteristics and task provenance to explicitly \texttt{control} generation attributes and implicitly \texttt{condition} generations at inference time. We fine-tune a base model to infer these markers automatically, which makes them optional at inference time. This principled and flexible approach yields pronounced improvements in performance, especially on examples from the long tail of the training distribution. While we observe an average lift of 5.7\% win rates in open-ended generation quality with our markers, we see over 9.1\% gains in underrepresented domains. 
We also observe relative lifts of up to 14.1\% on underrepresented tasks like CodeRepair and absolute improvements of 35.3\% on length instruction following evaluations.
}

\renewcommand{\FONTauthors}{\bfseries\Large}
\renewcommand{\FONTauthor}{\bfseries\Large}
\begin{document}

\begin{figure}[t]
    \centering
    \includegraphics[width=0.85\textwidth]{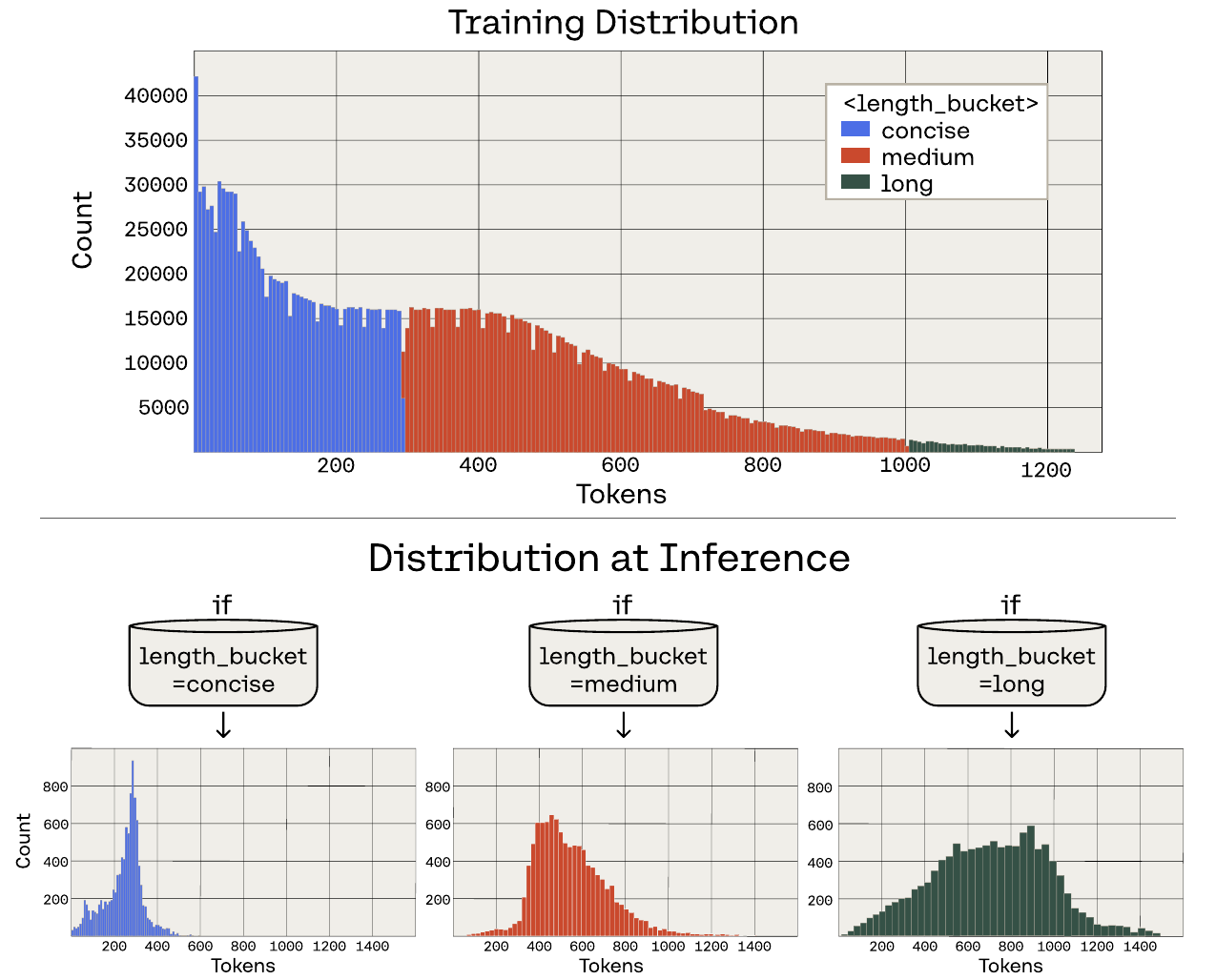}
    \caption{\textbf{Tapping into Distributions}: \textbf{(above)} illustrates the representation of various length buckets in the training distribution. \textbf{(below)} demonstrates the flexibility of the marker intervention on the mArena Hard test distribution. By modifying the \texttt{<length\_bucket>..</length\_bucket>} marker, the model can effectively tap into diverse training distributions, even for underrepresented length buckets. 
    }
    \label{fig:key_fig_lengthbucket}
\end{figure}

\section{Introduction}\label{sec:introduction}

Large language models (LLMs) are expected to perform well on many different tasks. Therefore, training data is a heterogeneous mix, where instances can vary greatly in terms of format, contents, tasks, and languages, e.g.\ code generation \citep{lozhkov2024starcoder, manh2023vault, kocetkov2022stack,zhong2017seq2sql} vs.\ MCQA \citep{singh2024aya, pal2022medmcqa}. At inference time, data points are not equally relevant, but it is often prohibitively expensive to go back and change the training distribution for each individual inference request. Hence, there is a mismatch in the distribution at training and inference time: training time distribution is often determined by ease of access to prior data collections and prior data augmentation efforts, while at inference time, new use cases might be underrepresented in the data but highly relevant to the user. 

To overcome this mismatch, techniques have been proposed to improve the conditioning of the output generation at inference. These involve prompt engineering \citep{wu-hu-2023-exploring,yu-etal-2023-exploring, wenjuan-etal-2024-prompt}, multi-shot examples \citep{brown2020language,lin-etal-2022-shot,winata-etal-2022-cross,logan-iv-etal-2022-cutting}, chain-of-thought \citep{wei2022chain,wang-etal-2023-plan,ranaldi-freitas-2024-aligning} or decoding strategies \citep{shi-etal-2024-thorough,snell2025scaling}. However, these approaches place an enormous burden on practitioners and developers to anticipate what strategies deliver the best performance. Furthermore, the effectiveness is dependent on the exact configuration for a particular model, e.g. the order of multi-shot examples plays a role~\citep{lu-etal-2022-fantastically}, and the wording of the prompt~\citep{anagnostidis2024susceptible}.
In this work, we ask \textit{Can we optimize our training protocols to both improve controllability by the user and improve performance on rare use cases at inference time?}

Our approach amounts to building a treasure map of hyper-detailed task-specific markers, to allow for real-time automatic targeting of long-tail features during inference. We note that some of the earliest generative models have used tags to improve performance. However, these often targeted a single feature at a time or were applied uniformly to an entire dataset. These early tags fell out of favor over the last few years, with the focus turning to prompt engineering for users to guide the generation themselves. However, there have been a few wider ecosystem changes which prompt (\textit{no pun intended}) revisiting the paradigm of adding markers to training, and also motivate this work: \textbf{1)} LLMs are now used by a far wider group of everyday users who can't be expected to be familiar with the complexities of prompt engineering, \textbf{2)} Many models are now served using an API which means training markers can be added automatically behind the API call (not visible to users), and hence can be far more complex and varied to guide and improve generations.

Our work is motivated by these two trends. We take a far wider view of training markers and explore a setting where a single data point can have up to 90 complex characteristics. We describe these as \texttt{Treasure Markers}, introduced at training time to provide a map to guide towards higher performance at inference time. We motivate that this approach is particularly beneficial for long-tail modeling. Our goal is that the treasure map approach is robust at test-time, so we also aggressively experiment with marker dropout during training. This is akin to asking the model to still find the treasure even with missing clues.

In this work, our primary contributions are as follows:
\begin{enumerate}
    \item \textbf{Introducing a more general framework for controllability:} We show that explicitly targeting controllability during training leads to pronounced gains at inference time, with little burden placed on the user. Training markers leads to significant downstream gains, ranging from a win rates increase on open-ended generations of 5.7\% on ArenaHard \citep{li2024crowdsourced} across the entire distribution of tasks relative to a model with no tags.  
    Our training marker framework offers remarkable flexibility, allowing for control over both aspects of form (output format, length) and semantic qualities (quality, tone, style) while also being completely \textit{optional} at inference, because the markers can be inferred accurately.

    \item \textbf{Long-tail lifts:} Training with explicit markers is an effective method for leveraging long-tail features at inference time, unlocking high-performance even for the distributions that are underrepresented in the training data. While our framework enables a relative improvement of 7.9\% on Code tasks over the baseline, we observe relative lifts of up to 14.1\% on tasks like CodeRepair, which are highly underrepresented in the training data.   
    \item  \textbf{Modeling underlying relationships:} We demonstrate that our approach effectively models underlying relationships in the data, as evidenced by a drastic reduction in length violation ($36.58\% \rightarrow 0.75\%$) in length-constraint instruction following, despite \textit{never} seeing a training sample with a prompt instruction designating length constraint. While significantly reducing length violation, the training markers also boost the generation quality with a 7.5\% relative gain ($14.36\% \rightarrow 21.85\%$) in win rates.
\end{enumerate}

\begin{figure}[t]
    \centering
    \begin{subfigure}{0.32\textwidth}
        \includegraphics[width=\linewidth]{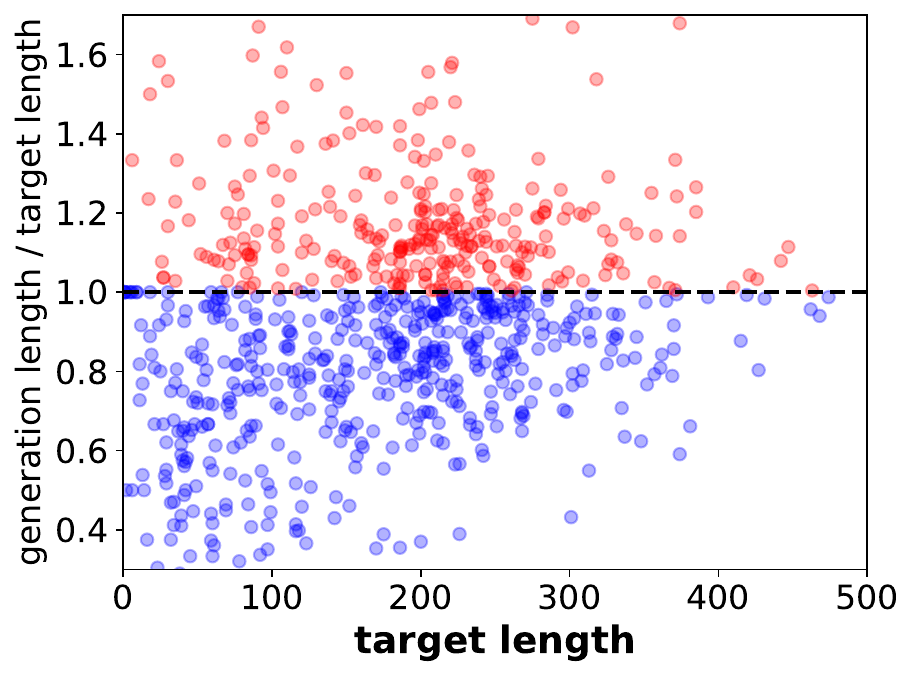}
        \caption{Baseline Model}
        \label{fig:li_orig_on_orig}
    \end{subfigure}
    \hfill
    \begin{subfigure}{0.32\textwidth}
        \includegraphics[width=\linewidth]{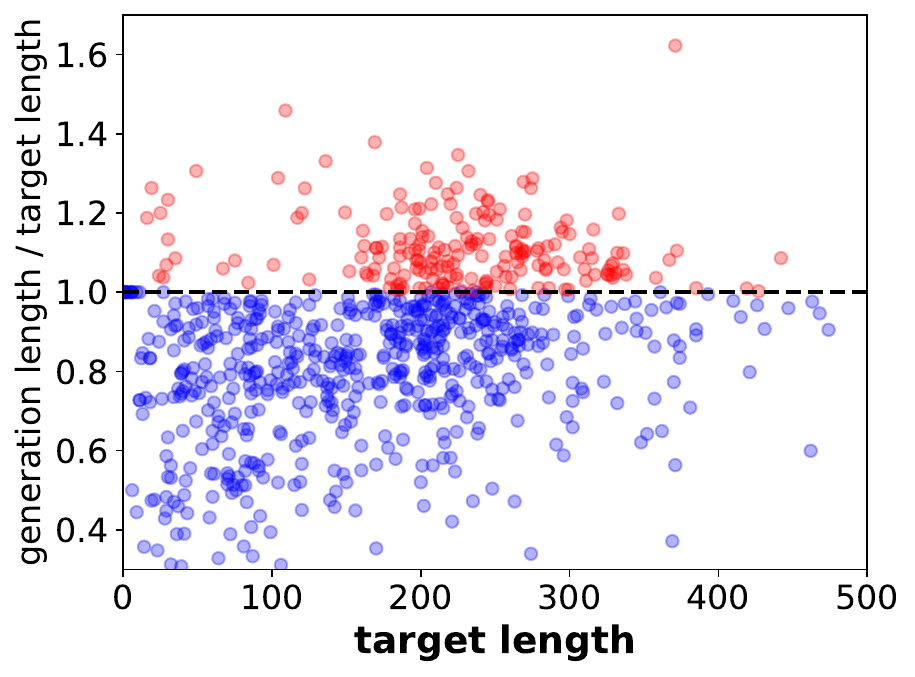}
        \caption{\textbf{TreasureMarked}}
        \label{fig:li_prefix_on_orig}
    \end{subfigure}
    \hfill
    \begin{subfigure}{0.32\textwidth}
        \includegraphics[width=\linewidth]{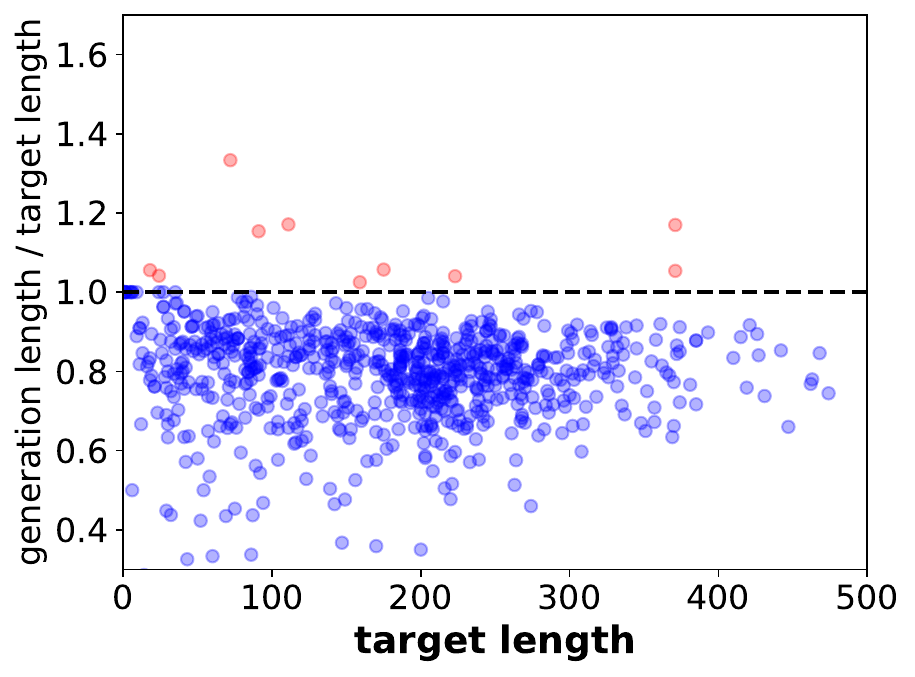}
        \caption{\textbf{TreasureMarked} (\textit{fixed})}
        \label{fig:li_prefix_on_prefix}
    \end{subfigure}
    \caption{\textbf{Modeling data features flexibly with training time markers:} Results on the length instruction following on the AlpacaEval-Length-Instructed(LI) dataset. While  \textbf{(a)} the \textit{baseline} violates the length constraint with \textbf{36.58\%},  \textbf{(b)} using the \textit{TreasureMarked} model and allowing to infer tags on the exact same dataset reduces the violation to \textbf{24.7\%}.  \textbf{(c)} Conveying the requirement via an explicitly inserted length marker to the prompts, the \textit{TreasureMarked} model violates the instruction on only \textbf{1.25\%} of the dataset.}
    \label{fig:li_key_fig}
\end{figure}

\section{Methodology}\label{sec:method}

\subsection{Overview of Training Time Markers}\label{sec:taxonomy}

We condition the output sequence $y$ given an instruction $x$ with added training markers $m$:
\begin{align}
    p(y|x,m) = \prod_{i=1}^{n} p(y_i \mid  x, m, y_{<i}).
\end{align}
These markers encompass several different attributes of the data, including estimated quality scores, domains, and languages (\ref{sec:markers}), which we store as a list of markers associated with a given data point (see \Cref{tab:format} for an example).

\begin{wraptable}{r}{0.4\textwidth}
\begin{tabular}{l}
\toprule
    \textcolor{blue}{\texttt{<MARKER\_LIST>}}\\
    \textcolor{blue}{\texttt{<domain>}} \textcolor{olive}{\textrm{Sciences}} \textcolor{blue}{\texttt{</domain>}}\\
    \textcolor{blue}{\texttt{<language>}} \textcolor{olive}{\textrm{French}}\textcolor{blue}{\texttt{</language>}}\\
    \textcolor{blue}{\texttt{/MARKER\_LIST}} \\
\bottomrule
\end{tabular}
\caption{\textbf{An example list of training time markers} formatted in a standardized template.}
\vspace{-0.5cm}
\label{tab:format}
\end{wraptable}
This template is treated as natural language and encoded with the same tokenizer as the text. We include the markers in both input (\textit{appended} to the prompt) and output space (\textit{prepended} to the completion), to induce the model to associate the properties of the generations with these characteristics. This reduces the burden on the practitioner or researcher at inference time, as the model learns to infer the correct markers. 

The finetuning objective thus becomes to minimize the negative log likelihood of the target generations including the template, given a prompt with an optional input template:
\begin{equation}
    -\frac{1}{|\mathcal{D}|} \sum_{d=1}^{|\mathcal{D}|} \log p_{\theta}(y_d,m_d \mid \text{dropout}(m_d), x_d)
\end{equation}

This approach ensures that the model learns to faithfully generate and adhere to the training markers when provided on the prompt side.

\textbf{Training markers dropout.}\label{sec:dropout} To avoid the model from becoming overly reliant on markers for completion or learning to trivially replicate the markers, we employ dual dropout strategies (dataset-level, sample-level) on the prompt space. In dataset-level dropout, we completely remove the training markers from the prompt for a random selection (\textit{defined as a percentage of the dataset}).
In sample-level dropout, we completely remove a random subset of training markers from each example (\textit{defined as a percentage of all markers associated with a given example}). To ensure the model consistently produces markers at inference time, we do not introduce dropout on the generation side.

\begin{figure}[t]
  \centering
  \begin{subfigure}[t]{0.56\textwidth} 
    \centering
        \includegraphics[width=\textwidth]{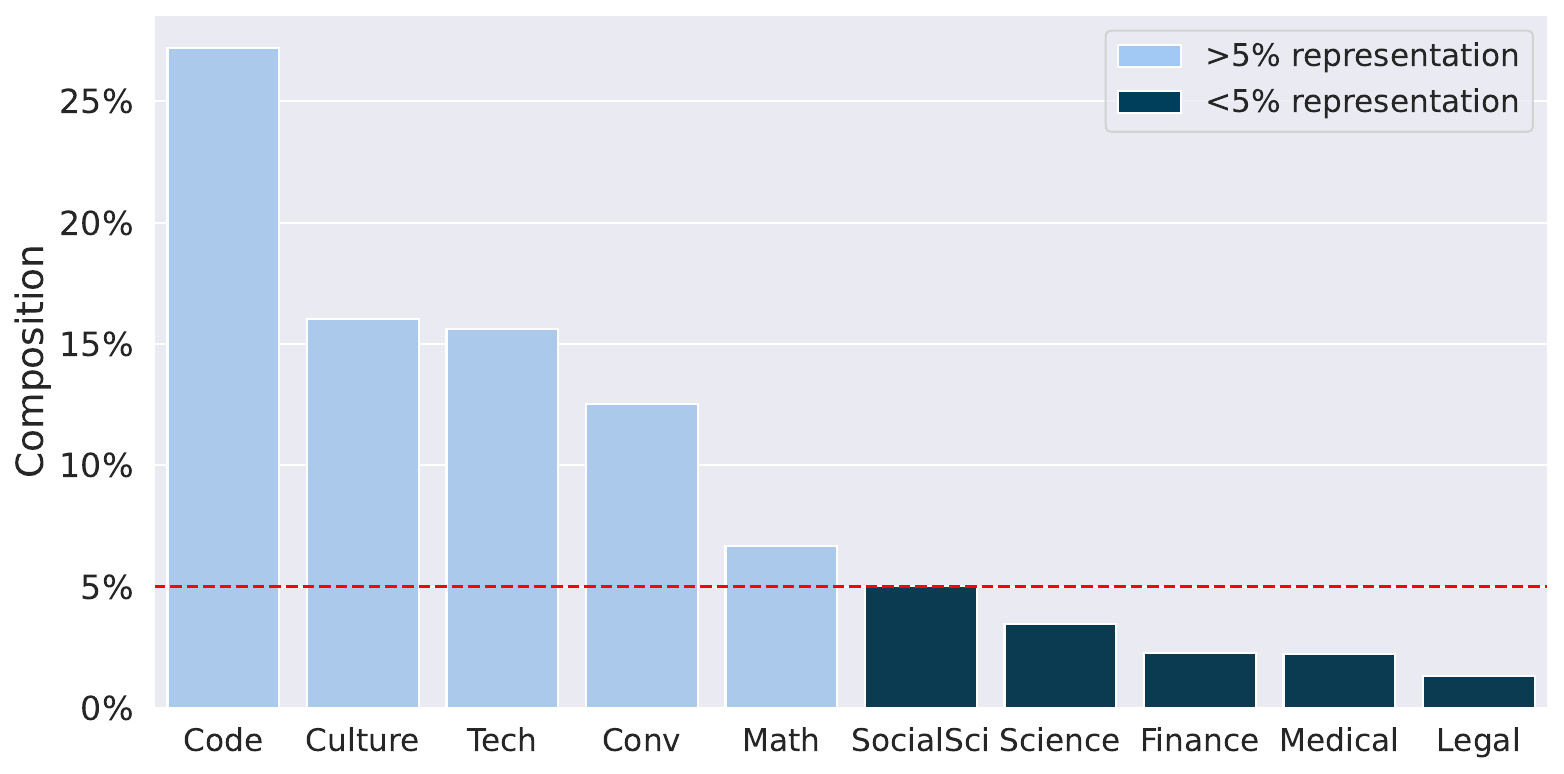}
    \label{fig:training_dist_domain}
  \end{subfigure}
  \begin{subfigure}[t]{0.38\textwidth} 
    \centering
    \includegraphics[width=\textwidth]{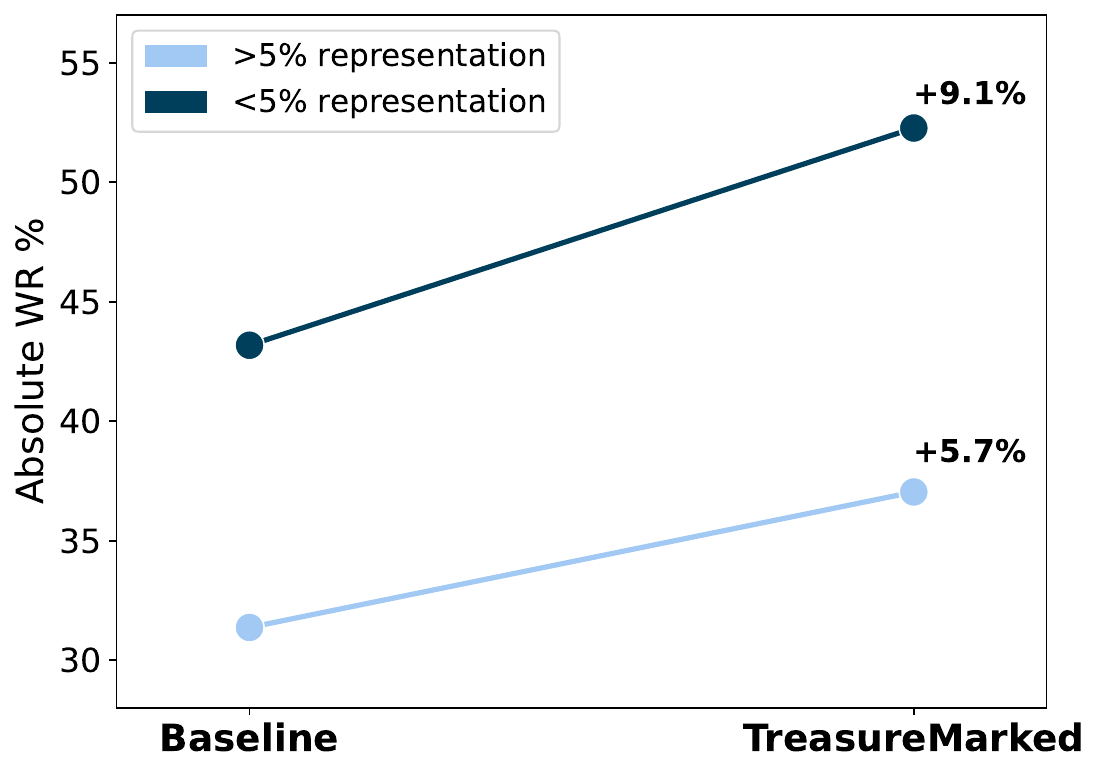}
    \label{fig:arenahard_wr_by_domain}
  \end{subfigure}
  \vspace{-0.2cm}
  \caption{\textbf{Long tail domains benefit more from training markers:} \textbf{(left)} Domain distribution in the training data used to fine-tune the model. \textbf{(right)} Improvements in win rates over the baseline against Gemma2-9B on both majority and minority subsets on Arena-Hard-Auto dataset \citep{li2024crowdsourced}. We group the test data into two sets of domains that have high (>5\%) and low (<~5\%) presence in training data.}
  \label{fig:gpt4o_wr_arenahard}
\end{figure}

\newpage
\section{Taxonomy for training time markers}
\label{app:taxonomy}
\begin{table*}
\centering
\begin{tabular}{p{3.1cm} p{4cm} p{6cm}}
\toprule
category & definition & values \\
\midrule
\texttt{<quality>} & Score indicating the quality assigned to a sample as annotated by a human or a RewardModel(RM).& 
\textit{float}\\
\hdashline
\noalign{\smallskip} 
\texttt{<quality\_bucket>} & \texttt{<quality>} bucketed into quartiles (calculated by language). 

1 indicating \textit{highest} quality and 4 indicating \textit{lowest} quality & \texttt{1,2,3,4}\\
\hdashline
\noalign{\smallskip} 
\texttt{<length\_tokens>} & \# of tokens 
& \textit{int}  \\
\hdashline
\noalign{\smallskip} 
\texttt{<length\_sent>} &  \# of sentences  & \textit{int} \\
\hdashline
\noalign{\smallskip} 
\texttt{<length\_para>} & \# of paragraphs & \textit{int} \\
\hdashline
\noalign{\smallskip} 
\texttt{<length\_bucket>} & \texttt{<length\_tokens>} bucketed by defined response length ranges & \texttt{concise, medium, long} \\
\hdashline
\noalign{\smallskip} 
\texttt{<task>} & Task-related information & \texttt{OpenEnded, Explanation, Translation, Classification, CreativeWriting, QuestionAnswering, InformationExtraction, Summarization, Rewrite, Reasoning, CodeGeneration, CodeFix, CodeTranslation, CodeExplanation} \\
\hdashline
\noalign{\smallskip} 
\texttt{<domain>} & Domain-related information & \texttt{Sciences, Technology, SocialSciences, Culture, Medical, Legal, Unspecified, Conversation, Code, Math} \\
\hdashline
\noalign{\smallskip} 
\texttt{<code\_type>} & \textit{(coding tasks)} Programming languages  & \texttt{python, javascript, cpp, cobol, java, go, rust, swift, csharp, php, typescript, shell, c, kotlin, ruby, haskell, sql} \\
\hdashline
\noalign{\smallskip} 
\texttt{<format>} & To model desired generation format & \texttt{MCQAnswer, ChainOfThought, XML, JSON, Enumeration, Tabular, Markdown, Latex} \\
\hdashline
\noalign{\smallskip} 
\texttt{<source>} & To model the source of the training data & \texttt{Human, Translation, Synthetic, Others} \\
\hdashline
\noalign{\smallskip} 
\texttt{<style>} & To model tone and style of the generation & \texttt{Formal, Informal, Custom} \\
\hdashline
\noalign{\smallskip} 
\texttt{<language>} & The language of the completion & The list of the languages are given in Section \ref{sec:lang-list}. \\
\bottomrule
\end{tabular}
\caption{\textbf{Comprehensive taxonomy for training time markers:} Our taxonomy contains 13 categories shown with their descriptions and values.}
\label{tab:taxonomy}
\end{table*}

\subsection{Taxonomy of Training Markers}\label{sec:markers}
We develop a comprehensive taxonomy around distinct groups of desired characteristics to capture key attributes of the training data, such as quality of the data, style, format, domain, and task. \Cref{tab:taxonomy} contains the taxonomy with definitions and the set of valid marker values.
We chose this selection of markers with inference-time use cases in mind: properties like quality, tone, style, and completion length are very desirable to control at inference time. We also focus on long-tail attributes with the goal of specifically targeting performance on underspecified parts of the distribution. To that end, we add hyperdetailed markers for task, domain and code type which tend to have highly skewed frequencies with some instances occurring far more frequently than others. 

To assign markers to samples in the training dataset, we utilize dataset-related information whenever possible and use an LLM to tag missing meta-information. Specifically, we use the multilingual open-weights model Command R+\footnote{Release blog of Command R+ \url{https://cohere.com/
blog/command-r-plus-microsoft-azure}} for tagging of markers for \texttt{<domain>}, \texttt{<task>}, \texttt{<format>} whenever unavailable from the dataset. To improve tagging performance, we use detailed definitions paired with few-shot examples to provide context for markers during annotation. We add markers across 23 languages, so we use in-language few-shot examples in each language.

Our extensive set of 90 unique markers fall into categories such as \textbf{Length}, \textbf{Style}, \textbf{Format}, \textbf{Quality}, \textbf{Source}, \textbf{Domain}, \textbf{Task}. We include an extensive description of all markers in \cref{app:taxonomy}. We describe the most frequently referenced categories below:
\begin{itemize}
    \item \textbf{Length}: are markers that allow for control of completion length. It includes a level of granularity ranging from \texttt{<length\_tokens>} and \texttt{<length\_sentences>} to broader categories such as \textit{concise}, \textit{medium}, and \textit{long}.

    \item \textbf{Language}: \texttt{<lang>} describes the language the completion is written in (i.e. Arabic, Japanese), enabling the model to improve language-specific generations and reduce language switching during inference. \texttt{<code\_type>} is specifically used to identify programming languages for coding-related tasks (i.e. python, c++). 
    \item \textbf{Quality}: \texttt{<quality>} provides a measurable score indicating the quality of a sample, often derived from human annotations or a Reward Model (RM). We also create a categorical marker \texttt{<quality\_bucket>} by using quartiles within language-specific subsets into \texttt{\{1,2,3,4\}}, offering a broader description of quality.
    \item \textbf{Domain}: overarching category of the knowledge required to answer a given prompt (i.e. Sciences, Technology, Medical). We annotate domain markers either using LLM tagging or derive from the source of the dataset for domains like Math and Code.   
    
    \item \textbf{Task}: \texttt{<task>} helps capture more fine-grained differences in task characteristics within a domain (i.e. summarization, reasoning, openended, explanation). Similar to the domain marker, we use LLM tagging or the data source information for obtaining task markers. 
\end{itemize}

\subsection{Experimental Set-up}\label{sec:experiments}

\textbf{Training with markers.} We use a 7-billion parameters proprietary base model which is pretrained using a data mixture that consists of texts from 23 languages covering half the worlds population. We train our base model on a training corpus containing 2.7M examples made up of our mixture of instruction-style data sources. 

\textbf{Training protocol.} Training for each variant spanned 8,000 steps, employed a cosine learning rate schedule with a warm-up phase, using a batch size of 32 and an evaluation batch size of 64. We train for 2 epochs with a peak learning rate of at 2.5 \(\times\) \(\displaystyle 10^{-4}\), achieved through 10 warm-up steps starting from a learning rate of 0.0, and then decay back to 1.25 \(\times\) \(\displaystyle 10^{-4}\). One fine-tuning run using 8,000 steps on 128 Nvidia H100 GPUs takes around 6 hours.

\textbf{Languages covered by the training markers.}\label{sec:lang-list} Our experiments cover 23 languages: \textit{Arabic, Chinese (simplified \& traditional), Czech, Dutch, English, French, German, Greek, Hebrew, Hindi, Indonesian, Italian, Japanese, Korean, Persian, Polish, Portuguese, Romanian, Russian, Spanish, Turkish, Ukrainian and Vietnamese}.

\textbf{Inference settings.} At inference time, we evaluate performance gains under two different settings. In the default setting, which we refer to as "\textbf{TreasureMarked}", we do not fix any of the markers at inference. This setting asks: \textit{Has the model learnt to infer the right markers without any intervention?} In the second setting which we refer to as "\textbf{TreasureMarked} (\textit{fixed})", we explicitly hardcode some of the markers at inference. This asks: \textit{if we manually set the value of some markers, can we drive gains in performance?} This is very reasonable for cases like quality, where we always want to steer model behavior towards higher quality generations.

\textbf{Baseline.} We compare both  "\textbf{TreasureMarked}" and "\textbf{TreasureMarked} (\textit{fixed})" against a model trained on the same data, but without added markers that we refer to as \textit{Baseline}. This allows for a clean comparison, and controls for the same amount of data seen in both variants.

\textbf{Core experimental variants and ablations.} In the next section, we evaluate a variety of ways a model trained with markers shines at inference time. We inspect three axes of control: \textbf{(1)} quality in \cref{sec:quality}, \textbf{(2)} length in \cref{sec:length}, and \textbf{(3)} language in \cref{sec:language}). Furthermore, we show how long-tail examples benefit from markers, even when only inferred at inference time (\cref{sec:domain}), specifically in coding tasks (\cref{sec:tail}) and for long generations (\cref{sec:length}). We present key experimental ablations, including understanding the impact of dropout applied to markers on downstream performance at inference time (\Cref{sec:dropout_ablation}).

\subsubsection{Evaluation}\label{sec:evaluation}

\textbf{Open-ended generation quality.} We evaluate the impact of markers on both the English Arena-Hard-Auto v0.1~\citep{li2024crowdsourced}, and a translated version of this dataset, m-Arena Hard \citep{dang2024aya} used for multilingual evaluation.
Arena-Hard-Auto is a challenging open-ended generation benchmark with prompts selected from user queries on Chatbot Arena. We measure Win Rate \% against our \textit{Baseline} model using GPT-4o.\footnote{We used gpt-4o-2024-05-13 as our judge model. Details: \url{https://platform.openai.com/docs/models/gpt-4o}} 

\textbf{Task-specific evaluations.} In addition, we evaluate the models on benchmarks specific to tasks such as code (generation, repair, translation) and length conditioned instruction following to narrow in on long-tail effects and controllability levers. We introduce each of these evaluations within the respective results sections. 

\textbf{Length evaluations.}\label{sec:len-eval} Given the original instruction in the AlpacaEval-LI dataset \citep{yuan2024following} contains the exact constraint, our \textbf{TreasureMarked} and \textbf{TreasureMarked} (\textit{fixed}) both contain explicit reference to the contraint. For  \textbf{TreasureMarked}, we present the original length-instructed prompt, allowing the model to deduce the associated tags. This approach evaluates the model's ability to extrapolate tags from instructions. in contrast, for \textbf{TreasureMarked} (\textit{fixed}), since the original instruction contains the exact constraint, we investigate an additional control strategy where we provide the constraint in the marker template if the taxonomy directly supports it. We remove the length instruction and append the corresponding \texttt{<length\_tokens>} tag with the appropriate value. \Cref{tab:prompt_examples} provides an example of an edited prompt. This strategy assesses the model's adherence to known templates and its ability to follow explicit length requirements that are only provided via the marker template. 

\section{Results}\label{sec:results}

\subsection{Impact of Treasure Markers on Open-Ended Generation}\label{sec:domain}

\textbf{Open-ended performance gains.} We measure Win Rates (\%) of the \textit{Baseline} and \textit{TreasureMarked} models against Gemma2-9B \citep{team2024gemma} as a common point of comparison, visualized in \Cref{fig:gpt4o_wr_arenahard}. We first consider our \textit{TreasureMarked} variant, markers are only included in training but are inferred from the model itself during inference. Overall, we obtain an absolute increase of 5.7\% in Win Rates from 32.1\% to 37.8\% across all tasks. This is reassuring, because it shows that markers at training time of the \textit{TreasureMarked} model can already make a positive change at inference time, \textit{even when only inferred by the model itself}, and even if the respective markers are rarely seen during training (e.g., for underrepresented domains).

\textbf{Performance on the long-tail.} One of our core hypotheses is that treasure markers will be particularly helpful at preserving or unearthing gains on the long-tail. To validate this hypothesis, we evaluate performance post-training on domains represented with different frequencies in the training-set. As seen in \Cref{fig:gpt4o_wr_arenahard}, \texttt{SocialScience, Sciences, Finance, Medical, and Legal} domains are particularly sparsely represented in the training data, each making up less than 5\% of the training data. In contrast, Code is best represented in the training dataset. With inferred treasure markers, while there is an improvement of +5.7\% across the higher-represented domains, we observe an even more pronounced gain of +9.1\% in the underrepresented domains. 

\subsubsection{Fixed Treasure Markers} 
\label{sec:quality}

We also explore adding explicit markers in \textit{TreasureMarked (fixed)}. Here, we specifically target quality and ask  \textit{Can we control the generation quality of the model as a latent feature, using training time markers?} To test this, we measure generation quality on m-Arena Hard \citep{dang2024aya} across 23 languages, by only adding markers related to quality. For each value \texttt{[1,2,3,4]} of \texttt{<quality\_bucket>}, we also include a \texttt{<quality>} score in conjunction with it. To obtain the \texttt{<quality>} score, we pick the 95\% percentile calculated language-wise from the samples in the training data from each respective bucket. 
As evaluation, we measure the generation quality by the same Reward Model used to score the data during training to compute win rates against the \textit{Baseline} model.

\begin{wrapfigure}{r}{0.45\textwidth}
    \centering
    \vspace{-0.5cm}
    \includegraphics[width=0.45\textwidth]{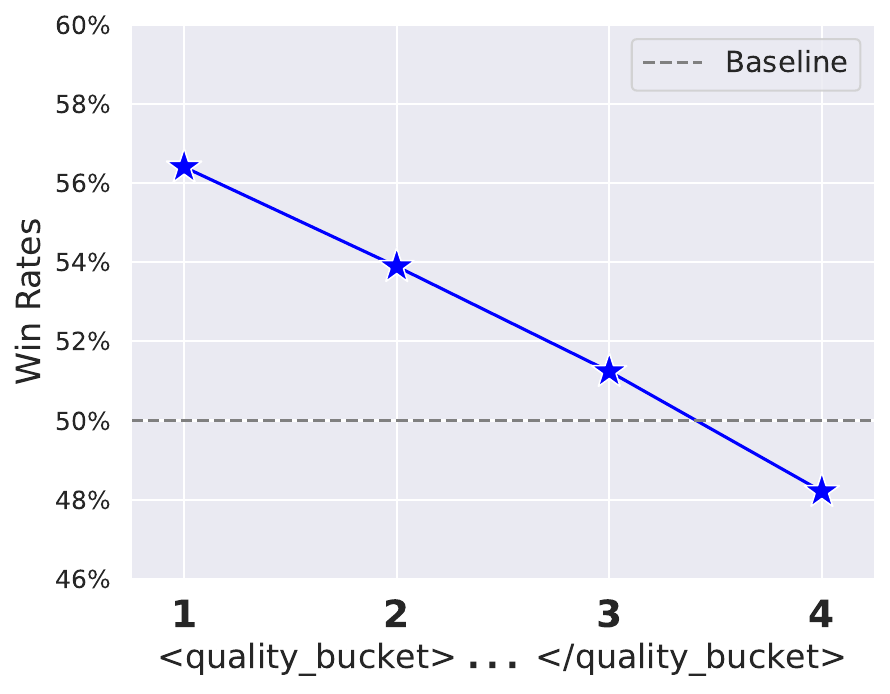}
    \caption{\textbf{Levers for Controlling Quality:} Changing the \texttt{<quality>, <quality\_bucket>} markers at inference time provides control over generation quality with Win Rates (as measured by internal Reward Model) going from $48.21\% \rightarrow 56.5\%$ over the \textit{Baseline} model, demonstrating successful control over quality as annotated in the training data.
    }
    \vspace{-0.8cm}
    \label{fig:quality_wr}
\end{wrapfigure}

Figure \ref{fig:quality_wr} demonstrates the amount of \texttt{control} introduced by training time markers with win rates under the RM going from $48.21\% \rightarrow 56.5\%$ just by changing \texttt{<quality>, <quality\_bucket>} at inference.  
These results showcase the potential of our framework, where markers representing a desired quality metric used during training yields control levers to leverage generations that tap into that quality metric at inference time.

\subsection{Impact of Treasure Markers on Targeted Performance of Specific Sub-tasks}\label{sec:tail}

\begin{figure*}[t]
    \centering
    \includegraphics[width=0.9\textwidth]{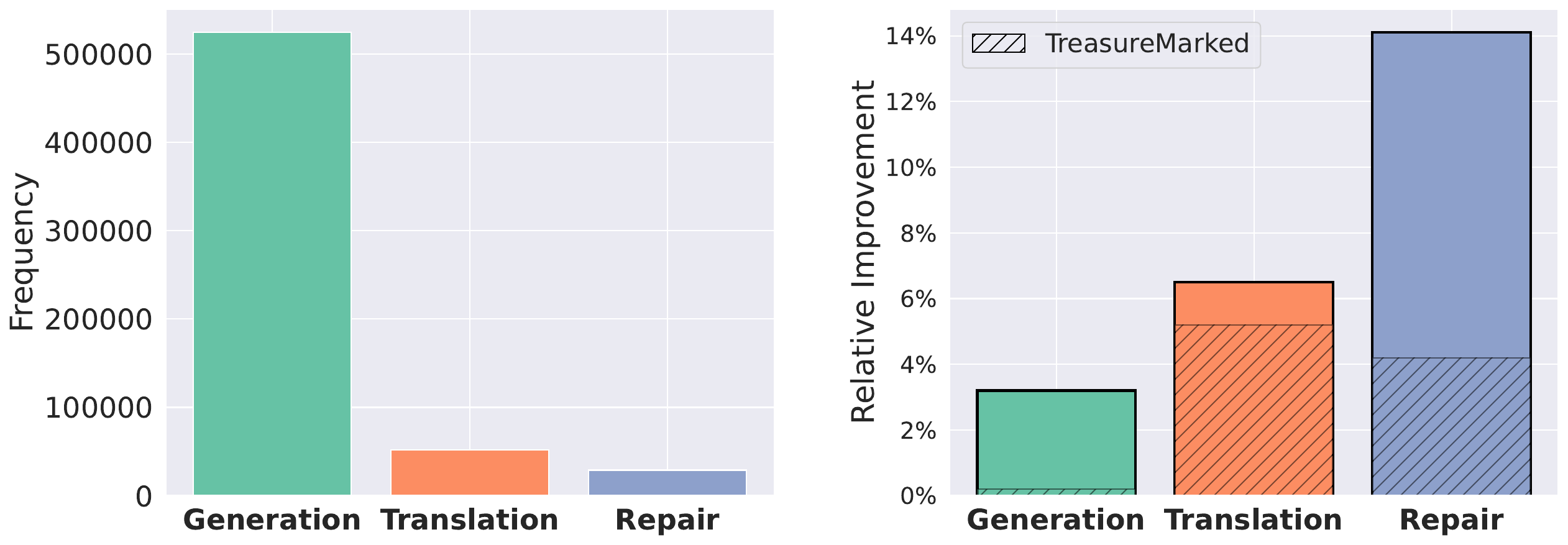}
    \caption{\textbf{Improvement on the Long Tail for Code tasks}: \textbf{(left)} Frequency of coding <task>s in the training dataset. \textbf{(right)} Despite being poorly represented in the training data, \textit{CodeRepair} achieves a 14.1\% relative improvement by leveraging targeted markers during inference further improving on the performance from the \textit{TreasureMarked} model with inferred markers.}
    \label{fig:code_plots}
\end{figure*}

\subsubsection{Code Performance}

For code, we evaluate our model on three tasks from HumanEvalPack~\citep{muennighoff2023octopack} dataset, and measure pass@1 rates. We use CodeSynthesis, CodeRepair, and CodeTranslation\footnote{The CodeTranslation task is created by an all-to-all mapping between the 6 languages in HumanEvalPack}, covering \textit{python, rust, java, javascript, go, c++}. 
These map to the following task markers in our taxonomy: \texttt{CodeGeneration}, \texttt{CodeFix}, and \texttt{CodeTranslation}.

During training, code comprises of 27.2\% of the overall training corpus. However, we specifically pick this domain because the distribution of coding subtasks differs significantly in frequency in the training corpus, as shown in \Cref{fig:code_plots}. \texttt{CodeRepair} and \texttt{CodeTranslation} are very rare coding subproblems, while \texttt{CodeGeneration} is heavily represented at 75.8\% within the coding data. 

\textbf{Long-tail gains.} We observe the largest gains on the long-tail code tasks. As seen in \Cref{fig:code_plots}, whether we provide the markers (\textit{TreasureMarked (fixed)}) or the model infers them, both rare coding problems (\texttt{CodeTranslation} and \texttt{CodeRepair}) show large lifts with up to 6.5\% and 14.1\% relative gain over the baseline respectively. We note that these gains are far higher than the gains observed for the far more frequent task of \texttt{CodeGeneration}, which only shows lifts of up to 3.2\% 
This shows that our framework benefits all parts of the distribution, but has disproportionate success enabling large lifts to highly infrequent features during training.

\begin{table}[]
    \centering
        \resizebox{0.8\textwidth}{!}{  

    \begin{tabular}{p{6.5cm} p{6.5cm}}
    
        \toprule
        \textbf{Original AlpcaEval-LI }& \textbf{TreasureMarked} \textit{(fixed)}\\
        \midrule
        \texttt{Answer the following instruction \textcolor{blue}{using} \textcolor{olive}{199} \textcolor{blue}{words or less.} }\newline
        \newline        
        \texttt{What are the names of some famous actors that started their careers on Broadway?} &  \texttt{What are the names of some famous actors that started their careers on Broadway?}\newline
        \textcolor{blue}{
        \texttt{<MARKER\_LIST>} \newline \texttt{<length\_tokens>}}\textcolor{olive}{199}\textcolor{blue}{\texttt{</length\_tokens>} \newline \texttt{</MARKER\_LIST>}}
        \\
        \bottomrule
    \end{tabular}%
    }
    \vspace{0.1cm}
    \caption{\textbf{Examples of length control strategies}: \textbf{(left)} Original instruction from AlpacaEval-LI dataset; \textbf{(right)} Modified instruction with constraint in the marker list.}
    \label{tab:prompt_examples}
\end{table}

\subsection{Length Control in Inference Time}\label{sec:length}

\begin{wraptable}{r}{0.44\textwidth}  
\centering
    \resizebox{0.44\textwidth}{!}{  
  \begin{tabular}{lcc}
    \midrule
    \textbf{Model} & \textbf{Violation} ($\downarrow$) & \textbf{Win Rates} ($\uparrow$) \\
    \hline
  \textit{Baseline} & 36.58\% & 14.36\% \\
  \textit{TreasureMarked} & 24.74\% & 19.48\% \\
  \textit{+(\textit{fixed})} & 1.25\% & 21.22\% \\
    \hline
  \end{tabular}%
  }
  \caption{\textbf{Length Instruction Following} \& generation quality on Alpaca-Eval LI.}
  \label{tab:li_metrics}
\end{wraptable}

To assess the impact of length conditioning during inference, we benchmark on the AlpacaEval-LI dataset \citep{yuan2024following}, which evaluates how faithfully LLMs adhere to length constraints. We complement the measurements for length violation with \textit{Win Rates (\%)} by evaluating valid samples against the dataset provided completions using GPT-4o.
We establish our baseline using completions generated by the \textit{Baseline} model.
Following a similar approach to \cite{yuan2024following}, we assess \textit{Violation (\%)} as the proportion of samples exceeding the specified length constraint (See Section  \ref{sec:len-eval} for details). 

\textbf{Improvements to length control.} In Table \ref{tab:li_metrics}, we show improvements of up to  35.3\% in length violation rates. This pronounced improvement results in a mere 1.25\% remaining violations for this evaluation set (essentially close to saturating performance on this evaluation). Even when the treasure markers are not explicitly provided but inferred directly by the model, we observe up to 11.8\% absolute decrease in violation rates. These improvements to instruction following are non-trivial, and also lead to overall win-rate gains of up to 6.86\%, ensuring quality is not compromised as length constraints are enforced.

\subsection{Machine Translation}\label{sec:machine_translation}

To study the effects of the markers on machine translation, we benchmark on WMT'24++\citep{deutsch2025wmt24++} and report translation performance from English to 22 languages ({$en \rightarrow xx$}) based on the languages seen in pretraining.
We use XCOMET-XL~\citep{colombo2023xcomet} for evaluation, a state-of-the-art machine translation evaluation metric~\citep{freitag-etal-2024-llms}.

Table \ref{tab:mt_metrics} shows the results with the relative improvement over the \textit{Baseline}. Training the model with markers and using them at inference time improves performance on 5 languages (\textit{es, id, it, pt, ro)} significantly with up to 1.18 point gains, while retaining performance on all other languages. This constitutes a remarkable improvement, especially given that the training data, up to the markers, is identical. According to the metric delta analysis in~\citep{kocmi-etal-2024-navigating}, improvements of such magnitudes are very likely to be confirmed in human evaluations. 

\begin{table}[t]
  \centering
  \resizebox{\linewidth}{!}{
  \begin{tabular}{cccccccccccc}
    \toprule
    en $\rightarrow$ xx & \textbf{ar} & \textbf{cs} & \textbf{de} & \textbf{el} & \textbf{es} & \textbf{fa} & \textbf{fr} & \textbf{he} & \textbf{hi} & \textbf{id} & \textbf{it} \\
    \midrule
    \textit{Baseline} & .6865 & .7485 & .8824 & .7463 & .8249 & .7099 & .789  & .7214 & .5158 & .776  & .8126 \\
    \textit{TreasureMarked} (\textit{fixed})    & .6844 & .755  & .8848 & .7500 & .8307 & .7072 & .7948 & .7166 & .5229 & .7874 & .8194 \\
                      & (-0.21) & (+0.65) & (+0.24) & (+0.37) & \textbf{(+0.58)} & (-0.27) & (+0.58) & (-0.48) & (+0.71) & \textbf{(+1.14)} & \textbf{(+0.68)} \\
    \midrule
    en $\rightarrow$ xx & \textbf{ja} & \textbf{ko} & \textbf{nl} & \textbf{pl} & \textbf{pt} & \textbf{ro} & \textbf{ru} & \textbf{tr} & \textbf{uk} & \textbf{vi} & \textbf{zh} \\
    \midrule
    \textit{Baseline} & .7368 & .7281 & .8103 & .7578 & .822  & .8048 & .7675 & .6669 & .7625 & .7593 & .7176 \\
    \textit{TreasureMarked} (\textit{fixed})    & .7342 & .7318 & .8117 & .7546 & .8281 & .8166 & .7627 & .6723 & .7575 & .756  & .7200 \\
    & (-0.26) & (+0.37) & (+0.14) & (-0.32) & \textbf{(+0.61)} & \textbf{(+1.18)} & (-0.48) & (+0.54) & (-0.50) & (-0.33) & (+0.24) \\
    \bottomrule
  \end{tabular}}
  \vspace{0.1cm}
  \caption{X-CometXL scores \citep{colombo2023xcomet} on WMT'24++ test sets \citep{deutsch2025wmt24++}. \textbf{Bold} differences are significant at $p \leq 0.05$ according to a paired T-Test and bootstrap resampling~\citep{koehn-2004-statistical} as implemented in \texttt{comet-compare}.}
  \label{tab:mt_metrics}
\end{table}

\begin{table}[ht]
  \centering
  \resizebox{\linewidth}{!}{
  \begin{tabular}{lccccccccccccccc}
    \toprule
    & \textbf{ar} & \textbf{de} & \textbf{es} & \textbf{fr} & \textbf{hi} & \textbf{id} & \textbf{it} & \textbf{ja} & \textbf{ko} & \textbf{pt} & \textbf{ru} & \textbf{tr} & \textbf{vi} & \textbf{zh} & \textbf{Avg.} \\
    \midrule
    \textit{Baseline} & 81.1 & 71.9 & 65.7 & 70.4 & 68.0 & 49.0 & 72.5 & 68.4 & 75.8 & 60.6 & 68.0 & 84.7 & 67.4 & 57.1 & 68.6 \\
    \textit{TreasureMarked (fixed)} & 88.4 & 84.4 & 82.7 & 79.8 & 73.7 & 66.7 & 82.8 & 83.8 & 85.0 & 72.2 & 86.6 & 78.6 & 82.8 & 62.6 & 79.58 ($\uparrow$ \textbf{10.98}) \\
    \bottomrule
  \end{tabular}
  }
  \vspace{0.1cm}
  \caption{Line-level pass rate on Complex Prompts from the Language Confusion Benchmark \citep{marchisio-etal-2024-understanding}.}
  \label{tab:lcb_metrics}
\end{table}

\subsection{Language Control in Inference Time}\label{sec:language}

As the final set of results, we focus on the effect of our training markers on ensuring a model responds in the language specified by the user. To evaluate this, we use the Language Confusion Benchmark \citep{marchisio-etal-2024-understanding} which measures the ability of a model to follow cross-lingual instructions such as ``Respond in French...'', to request completions in another language. We measure performance on the \textit{Complex Prompts} subset of the cross-lingual benchmark across 14 languages. Following \citep{marchisio-etal-2024-understanding}, we measure Line-level Pass Rate (\textbf{LPR}) that only deems a response "correct" if all lines in the generation match the user's desired language. During inference, we insert training markers present in the data into the prompt, but leave out the \texttt{<lang>} marker, since it is already present in the prompt. 

Table \ref{tab:lcb_metrics} shows results across 14 languages. Our model with training markers significantly improves language control performance in 13 out of 14 languages with an absolute gain of 10.98\% on average across 14 languages, showcasing a remarkable improvement in controllability of inference time. We observe the largest gains for Russian (+18.6\%) and the lowest gains for Chinese (+5.5\%). 

\newpage
\section{Key ablations and Discussion}

\begin{table}[]
    \centering
        \resizebox{0.8\textwidth}{!}{  

    \begin{tabular}{p{6.5cm} p{6.5cm}}
    
        \toprule
        \textbf{Original AlpcaEval-LI }& \textbf{TreasureMarked} \textit{(on-the-fly)}\\
        \midrule
        \texttt{Answer the following instruction using 199 words or less.}\newline
        \newline        
        \texttt{What are the names of some famous actors that started their careers on Broadway?} &  \texttt{Answer the following instruction using 199 words or less. \newline
        \newline
        What are the names of some famous actors that started their careers on Broadway?}\newline
        \textcolor{blue}{
        \texttt{<MARKER\_LIST>} 
        \newline \texttt{<domain>}}\textcolor{olive}{Culture}\textcolor{blue} {\texttt{</domain>} 
        \newline \texttt{<length_bucket>}}\textcolor{olive}{concise}\textcolor{blue} {\texttt{</length_bucket>} 
        \newline \texttt{<length_tokens>}}\textcolor{olive}{199}\textcolor{blue} {\texttt{</length_tokens>} 
        \newline \texttt{<task>}}\textcolor{olive}{QuestionAnswering}\textcolor{blue} {\texttt{</task>} 
        \newline \texttt{</MARKER\_LIST>}}
        \\
        \bottomrule
    \end{tabular}%
    }
    \vspace{0.1cm}
    \caption{\textbf{Examples of length control strategies}: \textbf{(left)} Original instruction from AlpacaEval-LI dataset; \textbf{(right)} Actual modified instruction by appending predicted markers annotated \textbf{on-the-fly} using Command-A}
    \label{tab:prompt_examples_onepass}
\end{table}

\subsection{Can markers be added on-the-fly at inference?}

Our framework of training-time markers provides significant flexibility for explicit control over generations at inference time. While users can manually insert these markers, another LLM can also automatically annotate prompts with training markers \textit{on-the-fly} before the generation step. In this section, to test the effectiveness of using another LLM to enrich an incoming prompt with the relevant markers at inference, we perform an ablation where we use Command A \citep{cohere2025command} as an annotator. At inference time, we make a single additional call to Command A to annotate a prompt with all the relevant markers using few-shot examples and then append them to the prompt. We use the AlpacaEval-LI evaluation, as it is an excellent test bed for this setup due to the existence of a clearly defined requirement in the prompt. \Cref{tab:prompt_examples_onepass} provides an example of one such annotation. The few-shot prompt used to annotate markers on-the-fly is provided in Appendix \ref{appendix:marker_annotation_onepass}. 

\begin{wraptable}{r}{0.52\textwidth}  
\centering
    \resizebox{0.5\textwidth}{!}{  
  \begin{tabular}{ccc}
    \toprule
    \textbf{Model} & \textbf{Violation} ($\downarrow$) & \textbf{Win Rates} ($\uparrow$) \\
    \midrule
\textit{Baseline} & 36.58\% & 14.36\% \\
\textit{TreasureMarked} & 24.74\% & 19.48\% \\
\textit{TreasureMarked (on-the-fly)} & 0.75\% & 21.85\% \\
    \bottomrule
  \end{tabular}}
  \caption{\textbf{\textbf{On-the-fly} control} (Alpaca-Eval LI): Using Command A to annotate markers at inference time drastically reduces violation rates to <1\% while improving Win Rates by +2.3\%}
  \label{tab:li_ablation_onepass}
\end{wraptable}

\Cref{tab:li_ablation_onepass} shows the results for this ablation. Similar to section \ref{sec:length}, we measure Violation (\%) and Win Rates (\%) for evaluation. When compared to using the \textit{TreasureMarked} model with the original prompts, we observe a drastic reduction in violation rates from 24.74\% to a mere 0.75\% with a 2.4\% relative improvement in Win Rates (from 19.48\% to 21.85\%). Compared to \textit{Baseline}, \textit{TreasureMarked (on-the-fly)} extends the gains and leads to a 35.8\% reduction in length violation and a 7.5\% improvement in Win Rates. These results demonstrate the potential gains possible by using an additional call at inference to annotate an incoming prompt with relevant markers using an external model.

\subsection{How do markers interact?} 

We perform an additional ablation on the AlpacaEval-LI dataset from  \cref{sec:length} to study the effect of adding more \textit{useful} markers at inference time. 
In addition to the \texttt{<length\_tokens>} marker that conveys the explicit length constraint, we annotate and add the \texttt{<domain>} marker, which we suspect carries implicit length biases (e.g. legal text might be longer than conversations), but should add helpful context to the prompt.  
With this we ask -- \textit{If multiple markers are added at inference, do their effects add up or cancel out?} 

\begin{wraptable}{r}{0.52\textwidth}  
\vspace{-0.5cm}
\centering
    \resizebox{0.5\textwidth}{!}{  
  \begin{tabular}{ccc}
    \toprule
    \textbf{Model} & \textbf{Violation} ($\downarrow$) & \textbf{Win Rates} ($\uparrow$) \\
    \midrule
\textit{TreasureMarked (fixed)} & 1.25\% & 21.22\% \\
+ \texttt{<domain>} & 1.87\% & 24.72\% \\
    \bottomrule
  \end{tabular}}
  \caption{\textbf{Multidimensional control} (Alpaca-Eval LI):
  Adding \texttt{<domain>} marker improves generation quality and hence Win Rates by +3.5\% working in conjunction with \texttt{<length\_tokens>}, without hurting the length control.}
  \label{tab:li_ablation}
\end{wraptable}

From \Cref{tab:li_ablation}, we observe that the effect of adding \texttt{<domain>} has a positive impact on the generation quality with a +3.5\% jump in win rates albeit at the cost of a slight increase in Violation Rate(\%). This indicates that there are multidimensional relationships that form between treasure markers during training and can be leveraged in conjunction to achieve desired characteristics at inference.

\subsection{What is the impact of the dropout on the marker prediction?}\label{sec:dropout_ablation}

\begin{wraptable}{r}{0.5\textwidth}  
\centering
    \resizebox{0.5\textwidth}{!}{  
  \begin{tabular}{ccccc}
    \toprule
    \textbf{dataset}\_\textbf{sample} & \textit{<domain>} & \textit{<task>} & \textit{<format>}  & \textit{<lang>}\\
    \midrule
  \texttt{0\_50} & 3.3\% & 1.9\% & 7.3\% & 1.2\% \\
  \texttt{50\_50} & 74.9\% & 53.6\% & 47.4\% & 99.2\% \\
  \texttt{70\_50} & 75.1\% & 51.4\% & 46.8\% & 99.1\% \\
    \bottomrule
  \end{tabular}}
  \caption{\textbf{Effect of dropout} on marker prediction. Using no dropout (dataset-level) prevents the model to learn predicting the correct marker across categories, hence, hurts the  flexibility of our framework. 
  }
  \label{tab:dropout_acc}
\end{wraptable}

To understand the impact of the marker dropout (\S~\ref{sec:dropout}), we train 
three variants with \textbf{dataset-level} dropouts of \texttt{[0\%, 50\%, 70\%]} while \textbf{sample-level} dropout is fixed to \texttt{50\%}. Our goal with dropout is to teach the model to infer markers without needing explicit guiding at inference time. However, too much dropout may impede the model from learning key patterns between tags and output properties. To evaluate this, we calculate the accuracy of the markers inferred by the model to the underlying markers assigned to m-Arena Hard and average across all 23 languages \citep{dang2024aya} 

In Table \ref{tab:dropout_acc}, we observe that the least extreme \textbf{dataset-level} dropout variant \texttt{0\_50} struggles to predict the correct marker at inference time. This is expected performance, since at training time, \texttt{0}\% dropout of markers across the dataset implies all training sample prompts have markers associated with it which makes it overly dependent on the presence of markers at inference time. At inference time, as this is not provided, accuracy is very low at 3.42\%. We note that at both  \texttt{50}\% and \texttt{70}\% dataset level dropout, we observe similar final abilities to infer the correct markers. Given this, unless specified elsewhere,   \texttt{50\%} \textbf{dataset-level} dropout is the default specification used throughout experiments since it strikes the best balance between learning and generalizaton.


\section{Related Work}
\textbf{From one- to multi-dimensional training data markers.} The idea of tagging inputs with markers in neural sequence modeling goes back to early applications in machine translation and language modeling. The motivation there was to leverage discrete features during training and inference to overcome data sparsity or imbalance and introduce levers of control. In early neural LMs, special tokens were added as markers to target a very specific attribute such as the topic~\citep{mikolov2012} or auxiliary features~\citep{aransa-etal-2015-improving} such as genre and length. In translation such markers were introduced to control attributes like the target language~\citep{johnson-etal-2017-googles} or desired output quality~\citep{caswell-etal-2019-tagged,riley-etal-2020-translationese,marie-etal-2020-tagged, larkin-etal-2021-like,freitag-etal-2022-natural} and text complexity~\citep{agrawal-carpuat-2019-controlling,marchisio-etal-2019-controlling}, but also language-specific nuances like politeness~\citep{sennrich-etal-2016-controlling,feely-etal-2019-controlling}, voice~\citep{yamagishi-etal-2016-controlling}, gender~\citep{kuczmarski2018},
domains~\citep{kobus-etal-2017-domain,britz-etal-2017-effective}, or diversity~\citep{shu-etal-2019-generating} of translations. Other works enriched the input representation during training with discrete linguistic features~\citep{sennrich-haddow-2016-linguistic} or document information~\citep{jehl-riezler-2018-document} for a better contextualization at inference time. 
Where and how tags should be placed best differ across applications~\citep{jehl-riezler-2018-document,wu-etal-2021-language}.

All of these were individual efforts that target one or two dimensions at a time, highly specialized for one trained target model and with training data for one particular task. Very limited work has been done on multidimensional markers~\citep{stergiadis-etal-2021-multi,ramnath-etal-2021-hintedbt}.
In contrast, our focus is on a much more general framework with a vast training corpus that targets general performance. Our approach is similarly general, where instead of a single feature, we want to enable a flexible approach that can be used for any text generation task. Furthermore, our goal is to explicitly target improving performance on the long-tail of underrepresented features. 

\textbf{From control in pretraining to control in instruction finetuning.} In LLM research, there are several related works that experiment with adding prefixes for \emph{control in pretraining}:
\citet{keskar2019ctrl} add control codes for desired text features in pretraining of a LLM derived from the structure of their source, i.e., subdomains or links of online texts and specific task labels for translation and QA. At inference time, values for these control codes are specific to steer the generation. 
\citet{gao2025metadata} further propose a cooldown schedule in pretraining going from marked data to unmarked data in order to not require prefixes at inference. 
\citet{yuan2024following} focus on length control by adding natural language length specification templates to \emph{fine-tuning} data for preference optimization.

In our work, we focus on the instruction finetuning stage and incorporate nuanced multi-dimensional markers (i.e. the user can specify length \emph{and} domain \emph{and} format). We circumvent a cooldown schedule by simply introducing marker dropout, hence requiring a much smaller volume of marked data at training time, and not a complete population of tags at inference time. With the option to fill markers on-the-fly, our framework is highly flexible and customizable.

\textbf{From encoded to inferred meta-information.} Related prefix and prompt tuning methods~\citep{li-liang-2021-prefix,lester-etal-2021-power} use continuous embeddings learned for special tokens representing markers in training to condition predictions for specific tasks at inference time.
\citet{shen2024tag} further break those into separate markers for domain and function. In our case, we directly embed prefixes with the same vocabulary as the LLM, smoothly integrating them into the sequence. In our experiments, we find that this helps format following even when specified in natural language and not markers (e.g. for desired output length and language \cref{sec:length,sec:language}). Attribute-based control in LLM generations has also been pursued with other methods, such as attribute classifiers ~\citep{Dathathri2020Plug} or learned attribute vectors~\citep{yang-etal-2023-tailor} --- see ~\citep{zhang2023survey} for a comprehensive survey.

\section{Conclusion}
In this work, we proposed adding markers to training data to map out potential ``treasures'' that can be retrieved at inference time, such as specific task configurations or quality characteristics. In our experiments on multilingual instruction-finetuning, we showed that these markers are a powerful tool to execute control (quality, length, output language) over generations, and at the same time have beneficial effects for generation quality of underrepresented portions of the training data, such as rare coding tasks. We found that dropout of training markers trains the model to infer missing markers at inference time. With this flexibility, we allow users to ``hunt treasures'' without having to tediously engineer prompts or few-shot examples for optimized performance.

\section{Acknowledgments}
We thank John Dang, Yiyang Nan, Thomas Euyang, Tom Kocmi, Tom Sherbone, Manoj Govindassamy, Cécile Robert-Michon, Leila Chan Currie, and other colleagues at Cohere and Cohere Labs for their support and thoughtful feedback.

\bibliography{main,anthology,addon}

\clearpage
\appendix

\section{Categories for Training Markers}\label{app:categories}

\textbf{Length}: 
\texttt{<length\_tokens>}, \texttt{<length\_sentences>}, and \texttt{<length\_paragraphs>} models granular control in generation length. We tokenize using language-specific Spacy models\citep{spacy2} to obtain token and sentence counts. For paragraph counts, we use the \texttt{"\textbackslash n\textbackslash n"} delimiter. \texttt{<length\_bucket>} categorizes generations into broader categories such as \textit{concise} (under 300 tokens), \textit{medium} (between 300 and 1,000 tokens) or \textit{long} (over 1,000 tokens), providing a more general level of control when needed.

\textbf{Format}: \texttt{<format>} is used to describe generations with specific output structures such as: JSON, Markdown, or tabular formats. This is particularly useful to condition stricter format requirements needed for real world use cases.

\textbf{Style}: \texttt{<style>} captures tone and manner of communication, distinguishing between different modes of expression such as "\textit{Formal}" and "\textit{Informal}". We also add a "\textit{Custom}" value to model for training examples where the user specifies a particular format. For instance, at inference if a user asks, "\texttt{Respond like Yoda you will}" this marker will allow the model to adapt its response to match the requested style. We annotate this marker by using dataset-related information.

\textbf{Language}: \texttt{<language>} describes the natural language of the generation, enabling us to model responses in specific languages during inference. We provide detailed markers across the 23 languages covered by our model. Our goal with this tag is to improve language-specific generations and reduce language switching where a prompt specified by a user in one language is not responded to in the same language in the completion. \texttt{<code\_type>} is specifically used to model programming languages for coding-related tasks. We annotate this marker by using dataset-related information.

\textbf{Quality}: \texttt{<quality>} provides a measurable score indicating the quality of a sample, often derived from human annotations or a Reward Model (RM). We utilize a proprietary reward model\footnote{ 
The RM is competitive with leading reward models on the RewardBench Leaderboard \citep{lambert2024rewardbench}(\url{https://huggingface.co/spaces/allenai/reward-bench})} to assign rewards to a subset of our training data. We also use these rewards to create a categorical marker \texttt{<quality\_bucket>} by using quartiles within language-specific subsets into \texttt{\{1,2,3,4\}}, offering a broader description of quality.
 
\textbf{Source}: \texttt{<source>} describes the origin of the data, distinguishing between human-generated content and other methods of data creation like synthetic and translation. We annotate this marker by using dataset-related information.

\textbf{Domain}: \texttt{<domain>} ensures that domain-specific knowledge is captured from training subsets, which can then be leveraged at inference to generate content that is relevant and accurate within a particular field. This is particularly crucial for inputs that could belong to multiple fields. For instance, when a user asks, "How do I calculate a factorial?", specifying the \texttt{<domain>} as either \texttt{Code} or \texttt{Math} provides valuable context for modeling the interaction. In cases where this marker cannot be obtained from the dataset information, we employ an LLM to annotate and provide our detailed prompt in \ref{appendix:llm_annotation}

\textbf{Task}: \texttt{<task>} defines the overall objective of the generation and helps capture task-specific behaviors, especially when outputs involve complex combinations of formats or actions. This marker is useful to model dataset-wise characteristics. We hypothesize this is particularly helpful for indicating  complex workflows during inference. For example, distinguishing between \texttt{Translation} and \texttt{CodeTranslation}, or \texttt{Rewrite} and \texttt{CodeFix}, enhances the descriptiveness of datapoints within the same training pool. In cases where this marker cannot be obtained from the dataset information, we employ an LLM to annotate and provide our detailed prompt in \ref{appendix:llm_annotation}

\section{LLM Annotation}\label{appendix:llm_annotation}
For the following training markers : \texttt{<domain>, <task>, <format>} we annotate using the multilingual open-weights model Command R+. 

We provide definitions and multilingual few-shot examples (except for \texttt{<format>}) to obtain high-quality annotations from the LLM. The prompt used for tagging is as follows: 

\subsubsection{\textbf{<domain>}}
\begin{Verbatim}[breaklines=true]

You are a helpful assistant whose goal is to classify the given prompt into a single class given the following definitions

`Sciences` : Topics related to the broad area of knowledge encompassing all scientific disciplines, including biology, chemistry, physics, earth sciences, and astronomy, which study the natural world through observation, experimentation, and analysis, aiming to understand fundamental principles and phenomena across various scales and aspects of the universe
`Technology` : Topics related to the broad area of knowledge encompassing all engineering and technical disciplines, including Computer Science, Software Engineering, Internet of Things(IoT), Cybersecurity, Data Science, Artificial Intelligence, Machine Learning and various engineering disciplines like Mechanical Engineering, Civil Engineering and Biotechnology
`SocialSciences` : Topics related to the broad area of knowledge encompassing all academic disciplines dedicated to the systematic study of human society, social relationships, and the structures that shape them, including fields like anthropology, economics, political science, psychology, and sociology, all focused on understanding how individuals and groups interact within a society and the factors influencing their behavior, cultural norms, and societal institutions
`Culture` : Topics related to the broad area of knowledge encompassing all cultural practices or beliefs within societies, including related concepts or behaviors that people within a culture group share and understand as belonging together, like food, art, language, family structure, societal norms or religious rituals
`Medical` : Topics related to the broad area of knowledge and practice encompassing all medicine and healthcare, including diagnosing and treating diseases, preventative measures, specialties like surgery, cardiology, oncology, pediatrics, and more, all built upon the foundation of basic medical sciences and patient care principles
`Finance` : Topics related to the broad area of knowledge encompassing activities like managing money, business ethics, investing, borrowing, lending, trading, budgeting, saving, and forecasting, essentially focusing on the acquisition, allocation, and management of capital within businesses
, individuals, and governments across various financial markets and instruments
`Legal` : Topics related to the broad area of knowledge encompassing Private, Public and Criminal Law, Criminal Justice, Law Enforcement, Policing, Justice Systems or Crime
`Conversation` : Topics related to Conversation, Chit-Chat or Roleplay
`Code` : Topics related to a specific subject/field within computer programming where software is designed and developed to solve problems related to a particular industry, business function, or area of expertise, essentially defining the target audience and unique requirements for the code being written including tasks like Code Generation, Code Fix and Code Explanation
`Math` : Topics related to the broad field of study that uses numbers, shapes, and formulas to describe and quantify the world, including areas like Logical Reasoning, Quantitative Calculation, Pattern Recognition, Formulating Conjectures, Arithmetic, Algebra, Geometry, Number Theory, Set Theory and Analysis

If you are unable to confidently assign one of the above classes, you will simply respond with `Unspecified` and nothing else.


Note:
- You are only to respond with the name of the class you believe best matches the domain of the example.
- You are only allowed to classify the example into one of the following tags :
[`Sciences`, `Technology`, `SocialSciences`, `Culture`, `Medical`, `Finance`, `Legal`, `Conversation`, `Code`, `Math`, `Unspecified`]

Here are a few examples : 

Prompt : What is photosynthesis?
Answer : `Sciences`

Prompt : What is the TCP/IP protocol and how does it work ? 
Answer : `Technology`

Prompt : How has globalization affected social cohesion ? 
Answer : `Social Sciences`

Prompt : What is an example of a popular dish that is available in multiple communities but known under different names? 
Answer : `Culture`

Prompt : How long does one have to fast before a fasting sugar blood test?
Answer : `Medical`

Prompt : Analyze the impact of microfinance initiatives on poverty alleviation in developing countries.
Answer : `Finance`

Prompt : What is the difference between a first-degree crime and a second-degree crime?
Answer : `Legal`

Prompt : Hey! How are you?
Answer : `Conversation`


Prompt : Given a variable x=3.142 in Python, how would I use an f-string to show just 1 decimal value?
Answer : `Code`

Prompt : Solve the quadratic equation: x² + 5x - 6 = 0
Answer : `Math`

Prompt : Use ABC notation to write a melody in the style of a folk tune.
Answer : 
\end{Verbatim}

\subsubsection{\textbf{<task>}}
\begin{Verbatim}[breaklines=true]
You are a helpful assistant whose goal is to classify the given prompt into a single class given the following definitions    

                
`CodeTranslation` : Tasks related to the process of converting source code from one programming language to another while preserving the code's functionality                                                                                                                                       
`CodeExplanation` : Tasks related to the specific process of explaining a snippet of code in a programming language                                                                                                                                                                                 
`CodeGeneration` : Tasks related to the specific process of generating a snippet of code in a programming language                                                                                                                                                                                  
`Explanation` : Tasks related to explaining a concept in any domain                                                                                                                                                                                                                                 
`CreativeWriting` : Tasks related to any form of writing that employs creative, literary or poetic techniques that displays imagination or invention including role-play                                                                                                                            
`QuestionAnswering` : Tasks related to any form of question answering, including open-ended questions, closed-ended questions about a given context and requests for information about a particular entity. This will also generally include your `what`, 'which', 'who', 'when' type questions     
`OpenEnded` : Tasks related to any form of open-ended text generation like chat, conversation or chit-chat                                                                                                                                                                                          
`InformationExtraction` : Tasks related to any form of information extraction usually involving some context                                                                                                                                                                                        
`Summarization` : Tasks related to any form of summarization including but not limited to abstractive summarization, extractive summarization or concise descriptions of content                                                                                                                    
`CodeFix` : Tasks related to the specific process of correcting/fixing a piece of code to achieve the desired result.                                                                                                                                                                               
`Reasoning` : Tasks involving any form of Ideation, Reasoning, Problem Solving, Instruction Following or Chain-of-Thought(CoT) in order to achieve the desired result.                                                                                                                              
`Rewrite` : Tasks involving any form of re-writing/re-phasing/re-wording/re-framing in order to achieve the desired result.                                                                                                                                                                         
`Classification` : Tasks related to specific request of classification where you are required to assign a thing to one of several groups                                                                                                                                                            
`Translation` : Tasks related to specific request of translating a given piece of text from one language to another language                                                                                                                                                                                
If you are unable to confidently assign one of the above classes, you will simply respond with `Unspecified` and nothing else.                                                                                                                                                                 
Note:                                                                                                                                                                                                                                                                                               
- You are only to respond with the name of the class you believe best matches the domain of the example.               
- You are only allowed to classify the example into one of the following tags :                                        
[`CodeTranslation`, `CodeExplanation`, `CodeGeneration`, `Explanation`, `CreativeWriting`, `QuestionAnswering`, `OpenEnded`, `InformationExtraction`, `Summarization`, `CodeFix`, `Reasoning`, `Rewrite`, `Classification`, `Translation`, `Unspecified`]                                           

Here are a few examples :                                                                                                 
Prompt : Translate the following Python function to equivalent JavaScript code that checks if a string is a palindrome.
            def is_palindrome(str):
                return str == str[::-1]
Answer : `CodeTranslation`

Prompt : Explain the following python function :
            def is_palindrome(str):
                return str == str[::-1]
Answer : `CodeExplanation`

Prompt : Generate a Python function to check whether a string is a palindrome.
Answer : `CodeGeneration`

Prompt : Explain briefly how the water cycle works
Answer : `Explanation`

Prompt : Translate the following sentence from English to Spanish, using a formal tone: 'We are pleased to announce the new partnership with our company.'
Answer : `Translation`

Prompt : You're a talk show host. Pick two guests that are wildly different from each other. Briefly introduce them
Answer : `CreativeWriting`

Prompt : Classify the sentiment of the following review as positive, negative, or neutral: 'The product exceeded my expectations!'
Answer : `Classification`

Prompt : What is the capital city of France?
Answer : `QuestionAnswering`

Prompt : Describe your ideal work environment
Answer : `OpenEnded`

Prompt : From the following news article, extract the names of the companies involved in the recent merger, along with the date the merger was announced.
Context:In a significant development in the tech industry, two leading companies have announced their merger, marking a new era of innovation and collaboration. The merger, which was officially announced on March 15, 2025, brings together TechInnovate Inc. and DigitalSolutions Corp, two giants in their respective fields. TechInnovate Inc, known for its cutting-edge research and development in artificial intelligence and machine learning, has been at the forefront of technological advancements. With a team of over 5,000 engineers and scientists, the company has consistently delivered groundbreaking solutions that have transformed various industries. DigitalSolutions Corp, on the other hand, is renowned for its expertise in software development and digital transformation. The company has a proven track record of helping businesses across the globe to modernize their
 operations and enhance their digital capabilities. With a workforce of over 10,000 professionals, DigitalSolutions Corp. has been a key player in driving digital innovation. The merger is expected to create a powerhouse in the tech industry, combining the strengths of both companies to offer comprehensive solutions that address the evolving needs of businesses and consumers. The combined entity will leverage TechInnovate's AI and machine learning capabilities with DigitalSolutions' software development expertise to develop next-generation technologies. Industry analysts predict that this merger will lead to significant advancements in areas such as autonomous systems, smart cities, and personalized healthcare. The synergy between the two companies is anticipated to drive innovation, improve efficiency, and create new opportunities for growth.
Answer : `InformationExtraction`

Prompt : Given the following story, provide a title that summarizes the idea behind the story:
Context:
There once was a girl who was frustrated with life and asked her father for advice. He asked her to bring an egg, two tea leaves, and a potato. He then started boiling water in three separate vessels. He put the egg, potato, and tea leaves in one vessel each. After a few minutes, he asked her to peel the egg and potato and strain the leaves. He explained to his daughter that:
The soft egg was now hard.
The hard potato was now soft.
The tea had changed the water itself.
When adversity is at our door, we can respond to it in different ways.

Moral: We decide how to respond to difficult situations.
Answer : `Summarization`

Prompt : Fix the code below to correctly identify a palindrome:
def is_palindrome(str):
        return str == str[-1]
Answer : `CodeFix`

Prompt : John has one pizza, cut into eight equal slices. John eats three slices, and his friend eats two slices. How many slices are left? Explain your reasoning step by step.
Answer : `Reasoning`

Prompt : Exaggerate this product description : 'Our new sneakers are comfortable, lightweight, and stylish.' to a paragraph that can be used by the marketing team
Answer : `Rewrite`

Prompt : Use ABC notation to write a melody in the style of a folk tune.
Answer : 
\end{Verbatim}

\subsubsection{\textbf{<format>}}
\begin{Verbatim}[breaklines=true]
You are a helpful assistant whose goal is to classify the given prompt into a single class given the following definitions


`MCQAnswer` : Tasks related to multiple-choice type question answering. These prompts will typically contain multiple choices provided either in bullet form or eumerated numerically/alphabetically. This also contains multiple-choice question answer generation tasks.
`ChainOfThought` : Tasks related to Chain-of-Thought(CoT) style question answering. This also contains CoT style question answer generation tasks
`Enumeration` : Tasks that involve enumeration, bullet points, lists or itemization of any form
`XML` : Tasks that involve XML generation, validation or processing in any form
`Tabular` : Tasks that involve table generation, validation or processing in any form
`JSON` : Tasks that involve JSON generation, validation or processing in any form
`Markdown` : Tasks that involve Markdown generation, validation or processing in any form

If you are unable to confidently assign one of the above classes, you will simply respond with `Unspecified` and nothing else.


Note:
- You are only to respond with the name of the class you believe best matches the domain of the example.
- You are only allowed to classify the example into one of the following tags :
[`MCQAnswer`, `ChainOfThought`, `Enumeration`, `XML`, `Tabular`, `JSON`, `Markdown`, `Unspecified`]

Prompt : Use ABC notation to write a melody in the style of a folk tune.
Answer : 
\end{Verbatim}


\section{Marker Annotation \textbf{on-the-fly}}\label{appendix:marker_annotation_onepass}
To annotate an incoming prompt with markers \textbf{on-the-fly} with Command A \citep{cohere2025command}, we use the following prompt 

\begin{Verbatim}[breaklines=true]
You are an expert tagger of information. You will be given a list of characteristics with their definitions and possible values.
You are to analyze the given prompt and only assign a value from a characteristic if it is absolutely applicable.
The format to be returned in XML format as specified below. Strictly follow this format and if no tags are applicable, simply return ```<MARKER_LIST></MARKER_LIST>``` 
You will only return one value from a characteristic(if applicable)

The characteristics are :
<length_tokens>
    Definition: any integer value denoting a requirement of an # of words or tokens in the prompt.
    Values: int
<length_sentences>
    Definition: any integer value denoting a requirement of an # of sentences or lines in the prompt.
    Values: int
<length_paragraphs>
    Definition: any integer value denoting a requirement of an # of paragraphs in the prompt.
    Values: int
<length_bucket>
    Definition: a grouping based on generation requirement of # of tokens. if # of tokens < 300, then 'concise'. If 300 < # of tokens < 1000, then 'medium'. If # of tokens > 1,000, then 'long'
    Values: list = ['concise', 'medium', 'long']
<task>
    Definition: to assign task-related information if evident from the prompt
    Values: list = ['OpenEnded', 'Explanation', 'Translation', 'Classification', 'CreativeWriting', 'QuestionAnswering', 'InformationExtraction', 'Summarization', 'Rewrite', 'Reasoning', 'CodeGeneration', 'CodeFix', 'CodeTranslation', 'CodeExplanation']
<domain>
    Definition: to assign domain-related information if evident from the prompt
    Values: list = ['Sciences', 'Technology', 'SocialSciences', 'Culture', 'Medical', 'Legal', 'Unspecified', 'Conversation', 'Code', 'Math']
<code_type>
    Definition: to specify the coding langugage. This is only applicable if the task is coding-related.
    Values: list = ['python', 'javascript', 'cpp', 'cobol', 'java', 'go', 'rust', 'swift', 'csharp', 'php', 'typescript', 'shell', 'c', 'kotlin', 'ruby', 'haskell', 'sql']
<format>
    Definition: to assign format-related information if specified from the prompt
    Values: list = ['MCQAnswer', 'ChainOfThought', 'XML', 'JSON', 'Enumeration', 'Tabular', 'Markdown', 'Latex']
<style>
    Definition: to assign style-related information if specified from the prompt. Use 'Custom' for customized styles if specified by the user.
    Values: list = ['Formal', 'Informal', 'Custom']
<lang>
    Definition: to assign language-related information if the generation language is specified from the prompt.
    Values: list = ['Arabic', 'Chinese', 'Czech', 'Dutch', 'English', 'French', 'German', 'Greek', 'Hebrew', 'Hindi', 'Indonesian', 'Italian', 'Japanese', 'Korean', 'Persian', 'Polish', 'Portuguese', 'Romanian', 'Russian', 'Spanish', 'Turkish', 'Ukrainian', 'Vietnamese']

You will only return a template and nothing else.

Here are some sample inputs and outputs:

prompt: "Give me a 4 paragraph summary of medical improvements that have occurred over the last two decades"
template:
```
<MARKER_LIST>
<domain>Medical</domain>
<lang>English</lang>
<length_bucket>medium</length_bucket>
<length_paragraphs>4</length_paragraphs>
<task>Summarization</task>
</MARKER_LIST>
```

prompt: "List the top 5 regions for food in Germany? Respond in German"
template:
```
<MARKER_LIST>
<domain>Culture</domain>
<format>Enumeration</format>
<lang>German</lang>
<length_bucket>concise</length_bucket>
<task>QuestionAnswering</task>
</MARKER_LIST>
```

prompt: "Answer the following instruction using 5 sentences or less.\n\nSolve this: 55+44+33+66"
template:
```
<MARKER_LIST>
<domain>Math</domain>
<length_bucket>concise</length_bucket>
<length_sentences>5</length_sentences>
<task>Reasoning</task>
</MARKER_LIST>
```

prompt: "Answer the following instruction
using 199 words or less.\n\nWhat are the names of some
famous actors that started their
careers on Broadway?"
template:
\end{Verbatim}

\end{document}